\PassOptionsToPackage{most}{tcolorbox}
\PassOptionsToPackage{table}{xcolor}
\PassOptionsToPackage{numbers,sort&compress}{natbib}
\documentclass[11pt]{article}

\usepackage[final]{arxiv}

\usepackage{times}
\usepackage{latexsym}

\usepackage[utf8]{inputenc}
\usepackage{amsmath,amssymb}
\usepackage{pifont}
\usepackage{placeins} 
\usepackage{booktabs}
\usepackage{multirow}
\usepackage{lipsum}
\usepackage{tabularx}
\usepackage{subcaption}
\usepackage{arydshln}
\usepackage{adjustbox} 
\usepackage{makecell}

\usepackage{siunitx}
\usepackage[T1]{fontenc}

\usepackage{inconsolata}
\usepackage{dblfloatfix}
\usepackage{graphicx}
\usepackage{enumitem}
\usepackage{cuted}
\usepackage{capt-of} 
\usepackage{graphicx}
\usepackage{dblfloatfix} 
\usepackage{caption}  
\usepackage{fontawesome5} 
\usepackage{hyperref}     

\usepackage{times}
\usepackage{latexsym}

\usepackage[utf8]{inputenc}
\usepackage{amsmath,amssymb}
\usepackage{pifont}
\usepackage{placeins} 
\usepackage{booktabs}
\usepackage{multirow}
\usepackage{lipsum}
\usepackage{tabularx}
\usepackage{subcaption}
\usepackage{arydshln}
\usepackage{adjustbox} 
\usepackage{makecell}

\usepackage{graphicx}
\usepackage{tikz}
\usetikzlibrary{fadings} 
\usepackage{wrapfig}
\usepackage{svg}
\usepackage{multirow}

\usepackage[utf8]{inputenc}
\usepackage[table]{xcolor} 
\usepackage{array}
\usepackage[most]{tcolorbox}
\usepackage{tcolorbox}
\usepackage{tabularx}
\usepackage{adjustbox}
\usepackage{makecell}
\usepackage{inconsolata}
\usepackage{dblfloatfix}
\usepackage{enumitem}
\usepackage{cuted}
\usepackage{capt-of} 
\usepackage{dblfloatfix} 
\usepackage{caption}  
\usepackage{fontawesome5} 
\usepackage{hyperref}     

\usepackage{graphicx}
\usepackage{booktabs}
\usepackage{multirow}
\usepackage[table]{xcolor}
\usepackage[most]{tcolorbox}
\definecolor{TableStripe}{RGB}{246,248,250}
\definecolor{TSRHighlight}{RGB}{232,244,255}
\definecolor{PromptBack}{RGB}{248,250,252}
\definecolor{PromptFrame}{RGB}{148,163,184}
\newenvironment{compactitemize}
  {\begin{itemize}
   \setlength{\itemsep}{1pt}
   \setlength{\parsep}{0pt}
   \setlength{\topsep}{2pt}
   \setlength{\partopsep}{0pt}}
  {\end{itemize}}
\newtcolorbox{promptbox}[1][]{
  enhanced,
  breakable,
  colback=PromptBack,
  colframe=PromptFrame,
  boxrule=0.5pt,
  arc=2pt,
  left=5pt,
  right=5pt,
  top=5pt,
  bottom=5pt,
  fonttitle=\bfseries\small,
  coltitle=black,
  #1
}

\newcommand{\methodname}{RoboSemanticBench}
\newcommand{\benchshort}{RSB}
\newcommand{\papertitle}{RoboSemanticBench:\\Diagnosing Semantic Grounding in Action Prediction for VLA Models}

\makeatletter
\AtBeginDocument{%
  \renewcommand{\@cite}[2]{\textsuperscript{[#1]}}%
}
\makeatother

\title{\papertitle{}}

\author{
  Bin Yu\textsuperscript{1,2,\thanks{Equal contribution}}
  Yao Zhang\textsuperscript{3,4,9,\footnotemark[1]}
  Haishan Liu\textsuperscript{2,\footnotemark[1]}
  Shijie Lian\textsuperscript{2,5,\footnotemark[1]}
  Yuliang Wei\textsuperscript{1,\thanks{Corresponding author}}
  \textbf{Xiaopeng Liu}\textsuperscript{3,6}\\
  \textbf{Zhaolong Shen\textsuperscript{2,7}}
  \textbf{Changti Wu\textsuperscript{2,8}}
  \textbf{Ruina Hu\textsuperscript{1,2}}
  \textbf{Bailing Wang\textsuperscript{2,3}}
  \textbf{Cong Huang\textsuperscript{2,3}}
  \textbf{Kai Chen\textsuperscript{2,3,9,\footnotemark[2]}}
  \\[2ex]
  \textsuperscript{1}HIT\quad
  \textsuperscript{2}ZGCA\quad
  \textsuperscript{3}ZGCI\quad
  \textsuperscript{4}WHU\quad
  \textsuperscript{5}HUST\quad
  \textsuperscript{6}HKUST(GZ)\quad
  \textsuperscript{7}BUAA\quad
  \textsuperscript{8}ECNU\quad
  \textsuperscript{9}DeepCybo
}

\begin{document}
\maketitle

\begin{center}
    \vspace{-50pt}
    \faGithub\hspace{6pt}\href{https://github.com/ZGC-EmbodyAI/RoboSemanticBench}{\texttt{\color{black}https://github.com/ZGC-EmbodyAI/RoboSemanticBench}}
\end{center}

\vspace{5pt}

\let\thefootnote\relax\footnotetext{Work done at Zhongguancun Academy (Beijing).}

\begin{abstract}
Vision-language-action (VLA) models are built on the premise that semantic understanding from pretrained language or vision-language backbones should guide robot action prediction. Yet robot fine-tuning is optimized as imitation over task-specific action distributions, and many evaluations can be solved through visual or instruction-action shortcuts. We introduce \textbf{RoboSemanticBench (RSB)}, an embodied benchmark for diagnosing \emph{semantic grounding in action prediction}: whether post-trained VLA models can use complex instruction semantics to select and manipulate the correct physical target. In each episode, a robot receives a multiple-choice math or general-knowledge question, observes candidate answer blocks, and must grasp the block corresponding to the correct answer. RSB covers controlled arithmetic, grade-school mathematical understanding, and commonsense or factual understanding under four-choice and ten-choice suites. Across representative VLA models, we find that many policies learn to grasp candidate blocks but select the semantically correct block at near-random or below-random rates after controlling for grasp success, revealing a persistent gap between backbone-level semantic competence and action prediction.
\end{abstract}

\section{Introduction}
\label{sec:intro}

\begin{figure}[t]
  \centering
  \includegraphics[width=0.40\textwidth]{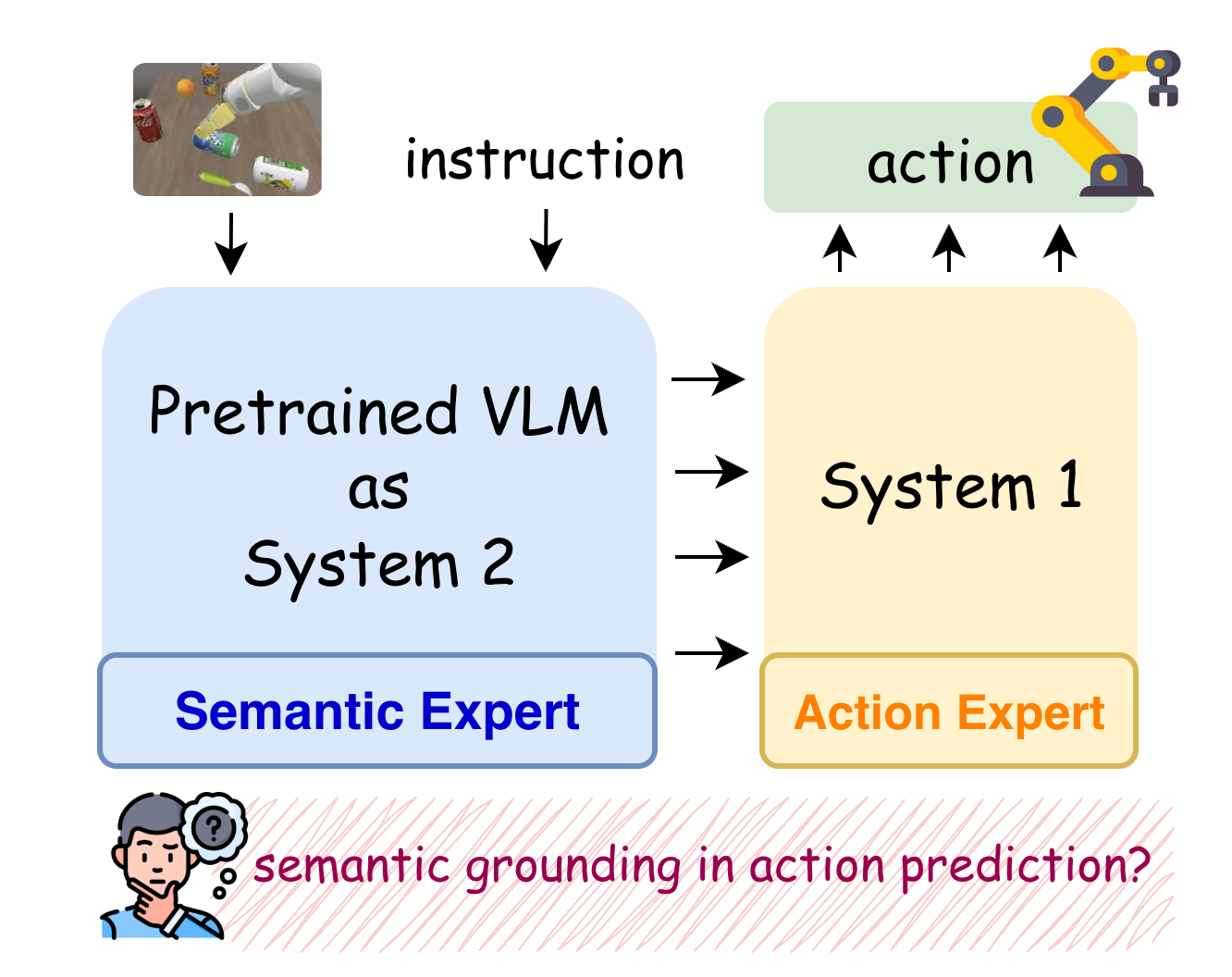}
  \caption{Dual-system VLA architecture and the semantically grounded action prediction question. \methodname{} tests whether the System-2 Semantic Expert's competence is preserved and used by the System-1 Action Expert after VLA post-training.}
  \label{fig:intro}
\end{figure}

Vision-language-action (VLA) models are motivated by a compelling promise: the semantic competence of pretrained language or vision-language models should become part of robot action prediction. Representative systems, including $\pi_0$, are often described as dual-system architectures with a low-frequency System-2 \emph{Semantic Expert} and a high-frequency System-1 \emph{Action Expert}~\cite{OpenVLA_24,PI0,ChatVLA_25}. As illustrated in Figure~\ref{fig:intro}, the Semantic Expert processes observations and instructions, while the Action Expert turns its outputs and proprioception into continuous actions. Under this view, a VLM backbone should not merely condition a controller on text; its general knowledge and semantic understanding should participate in the action-generation pathway.

The central concern is that this promise may not survive current VLA post-training pipelines. Robot demonstrations are much smaller and more task-specific than language pretraining corpora, and imitation losses can reward fitting conditional action distributions even when the Semantic Expert is weakened or decoupled from the Action Expert. In a typical demonstration dataset, an instruction is paired with a successful trajectory, but the loss rarely forces the model to expose the semantic decision that made the trajectory correct. The policy may therefore learn instruction-action or visual shortcuts rather than use language semantics to determine which action is correct, creating a direct tension with the motivation for VLA models~\cite{VLM2VLA_25,VLM4VLA_25,HowInherit_25}.

This tension matters because realistic robot instructions are often complex, underspecified, or require commonsense interpretation, whereas many simulation benchmarks use short explicit commands such as picking a named object or moving to a visible location. In such settings, VLA policies can ignore language and still score well by exploiting visual shortcuts or dataset regularities~\citep{LangForce,CAG_26_arxiv,PriorVLA_26_arxiv}. A high task success rate can therefore be ambiguous: it may indicate genuine instruction understanding, or merely that the benchmark admits non-semantic shortcuts. Recent studies further suggest that preserving the general semantic competence of pretrained VLM backbones is important for action-generation generalization~\citep{TwinBrainVLA,UAM_26_arxiv,PriorVLA_26_arxiv}. Yet existing benchmarks rarely measure whether such competence is actually grounded in action prediction. We call the missing capability \underline{\emph{\textbf{semantic grounding in action prediction}}}: when controlling a robot, a VLA model should understand the semantics of a human instruction, ground them in the current observation, and act according to the user's intent.

\begin{figure*}[t]
  \centering
  \includegraphics[width=\textwidth]{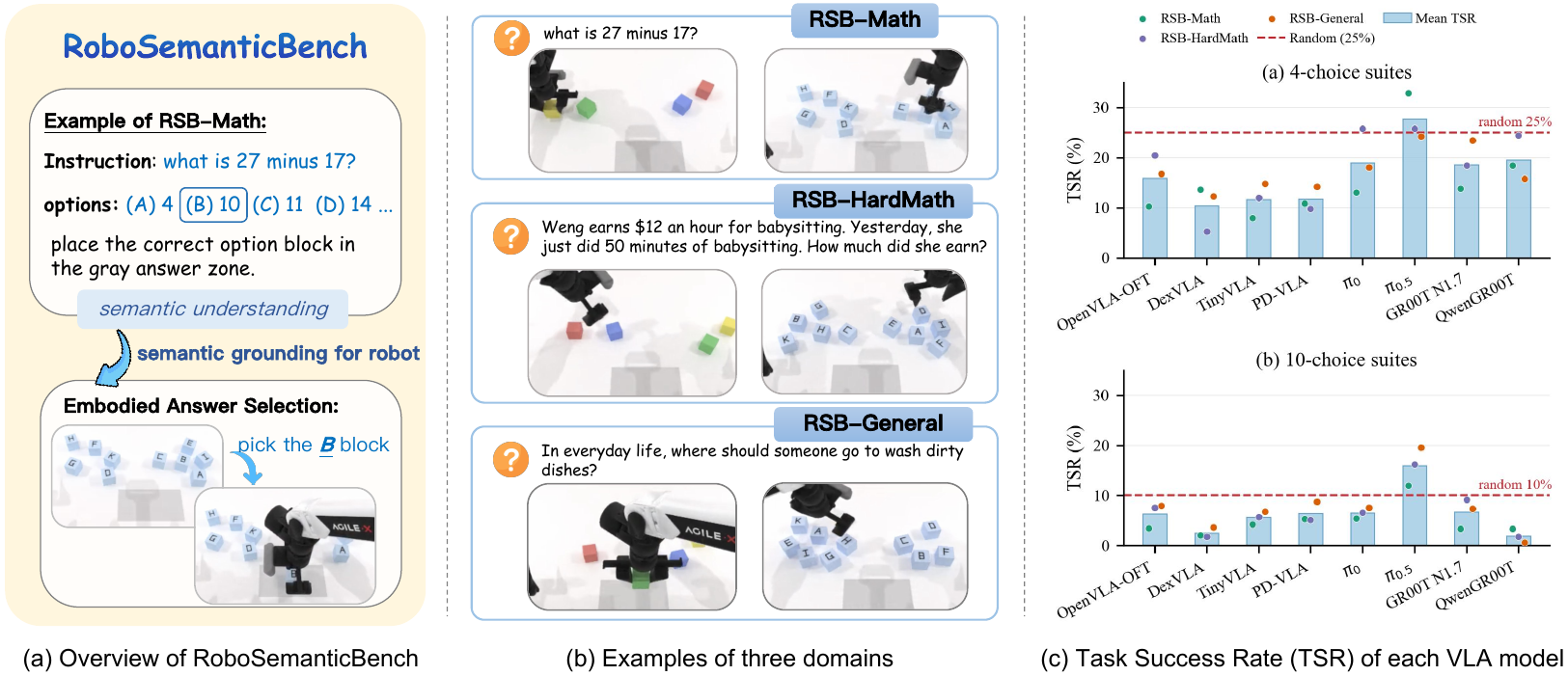}
  \caption{Overview of \textbf{\methodname{}} (\textbf{RSB}). (a) RSB turns a multiple-choice semantic question into an embodied answer-selection task: the VLA must understand the instruction, identify the correct option, bind it to the corresponding visible target, and execute the grasp. (b) The benchmark covers three semantic domains, RSB-Math, RSB-HardMath, and RSB-General, under both 4-choice and 10-choice settings. (c) Representative VLA models achieve low Task Success Rate (TSR) and often stay near or below the random-selection baseline, even though the manipulation primitive is simple. This highlights the semantic grounding failure in action prediction exposed by RSB.}
  \label{fig:rsb_overview}
\end{figure*}

We propose \textbf{\methodname{}} (\textbf{\benchshort{}}), a controlled diagnostic for semantic grounding in action prediction. As shown in Figure~\ref{fig:rsb_overview}, each episode asks a VLA to understand a multiple-choice math or general-knowledge instruction, bind the correct answer to a visible target, and execute the corresponding grasp. Because the manipulation primitive is fixed while the semantic content and option set vary, \benchshort{} tests whether instruction semantics guide target selection rather than merely fitting instruction-action correlations.

This design is deliberately simple in the motor domain but demanding in the semantic domain. It exposes two questions: whether the Semantic Expert can still solve the instruction-level problem, and whether the Action Expert can follow the resulting implicit target during action prediction. A model that can reliably grasp candidate targets should still fail if it cannot ground the instruction's semantic answer into the target-selection action. We therefore compare grasping any candidate block with grasping the correct one: a large GSR--TSR gap reveals successful manipulation but failed semantic grounding. This makes \benchshort{} a diagnostic of the VLA interface between semantic understanding and action prediction, rather than a generic test of grasping skill.

Our contributions are:
\begin{compactitemize}
    \item We introduce \textbf{\methodname{}}, a benchmark that turns math, hard-math, and general-semantic understanding into embodied answer selection across six evaluation suites.
    \item We define GSR, TSR, and nSG to separate low-level grasping from semantic target selection and diagnose whether instruction semantics participate in action prediction.
    \item We evaluate representative VLA models and show that many perform near or below random semantic target selection after controlling for grasp success.
    \item We report negative interventions and error analyses, showing that CoT-style ReasoningVLA and VLA cotraining do not reliably close the semantic grounding gap.
\end{compactitemize}

\section{Related Work}

\subsection{Vision-Language-Action Models and Benchmarks}

Recent VLA models connect pretrained vision-language representations with robot control. Early embodied foundation models show that web-scale visual-language pretraining can provide reusable semantic knowledge for robot tasks~\citep{PaLM-E_23_arxiv,RT2_23_arxiv}; later systems scale this idea through large robot datasets and open generalist policies, including Open X-Embodiment/RT-X~\citep{OXE_24}, Octo~\citep{Octo_2024}, OpenVLA~\citep{OpenVLA_24}, $\pi_0$~\citep{PI0}, $\pi_{0.5}$~\citep{PI05_25}, and GR00T N1~\citep{GR00T_25}. Recent work further studies how to adapt VLM backbones into VLA policies without losing pretrained capability~\citep{VLM2VLA_25,TwinBrainVLA,UAM_26_arxiv,PriorVLA_26_arxiv}. Yet standard task success does not reveal whether semantic competence is actually preserved and used for action prediction.

Benchmarks for language-conditioned manipulation measure complementary abilities. CALVIN and LIBERO study multi-task and lifelong manipulation~\citep{CALVIN_22,libero}, SimplerEnv evaluates policies through simulation-based reproduction~\citep{SimplerEnv_24}, and other suites scale household tasks, robot data generation, world-knowledge manipulation, or VLA comparison~\citep{RoboCasa_24,VLABench_24_arxiv,RoboTwin2_25,VLA-Arena_25,VLA-Eval_26}. These benchmarks are valuable for robustness and generalization, but their task success metrics often entangle motor execution, object recognition, and language grounding. \methodname{} instead fixes the manipulation primitive and varies the instruction's semantic content, making it a targeted diagnostic for whether semantic understanding is grounded in robot action prediction.

\subsection{Diagnosing Language Grounding and Shortcut Behavior in VLAs}

Several studies question whether VLA policies use language as intended. Policies may exploit visual regularities or action priors, ignore instructions under conflicting visual evidence, or suffer degraded VLM representations after robot fine-tuning~\citep{TwinBrainVLA,CAG_26_arxiv,PriorVLA_26_arxiv,VLM2VLA_25,VLM4VLA_25,HowInherit_25}. Related diagnostics probe instruction perturbations, counterfactual commands, linguistic diversity, and distribution shift~\citep{LangGap_26,FromIntentionToExecution_25,AttentionRecalibration_26,limited_diversity}. \methodname{} complements them by making the instruction itself a semantic problem: the policy must answer a math or general-knowledge question, bind the answer to an observed target, and execute the grasp. This evaluates \emph{semantic grounding in action prediction} rather than language sensitivity alone.

\section{RoboSemanticBench}
\label{sec:method}

\textbf{Motivation.}
Most embodied benchmarks are designed to evaluate manipulation robustness, task generalization, or policy transfer, but their language instructions are often semantically simple. For example, LIBERO, SimplerEnv, and RoboTwin benchmarks commonly construct evaluation instructions from predefined language templates, such as short object-selection or placement commands~\citep{libero,SimplerEnv_24,RoboTwin2_25}. These instructions are useful for controlled manipulation evaluation, but they cover a narrow semantic range and rarely require deeper commonsense, arithmetic, or problem-solving semantics. As a result, they do not fully reflect the architectural promise of the VLA paradigm, where a pretrained Semantic Expert is expected to contribute rich language understanding to action generation, and they differ substantially from realistic human instructions that are often diverse, underspecified, and semantically loaded. \methodname{} is designed to fill this diagnostic gap by making semantic understanding a necessary condition for selecting the correct physical action.

\subsection{Benchmark Overview}
\label{sec:method-overview}

\methodname{} evaluates whether semantic decisions made from language are used to generate robot actions. Each episode contains a question $q$, candidate options $\mathcal{O}=\{o_1,\ldots,o_N\}$, visible answer blocks $\mathcal{B}=\{b_1,\ldots,b_N\}$, and an option-to-block mapping $m:\mathcal{O}\rightarrow\mathcal{B}$. The policy receives an instruction containing $q$ and the mapping, observes the scene, and succeeds only if it moves the block associated with the correct answer into the answer zone.

The physical action is always a single answer-selection primitive: pick the selected candidate block and place it in the gray answer zone. The semantic content changes across episodes, requiring the policy to solve the question, compare options, ground the chosen option to the visible block, and execute the pick-and-place action. Importantly, the correct target is not tied to a fixed color, letter, position, or trajectory; it is determined by the question and the episode-specific option mapping. This isolates whether instruction semantics participate in action prediction.

\subsection{Semantic Task Construction}
\label{sec:semantic-tasks}

\benchshort{} contains three semantic subsets with different semantic demands.

\textbf{(i) RSB-Math} uses procedurally generated arithmetic questions. Each question samples one of three controlled forms: two-digit addition, two-digit subtraction, or one-digit by two-digit multiplication. The correct numerical answer is mixed with nearby distractors, so the policy must compute the result rather than rely on superficial option patterns.

\textbf{(ii) RSB-HardMath} uses grade-school word problems derived from GSM8K \cite{GSM8K_21}. The problem text is used as $q$, the final answer is the correct option, and distractors are taken from prepared option fields or generated around the answer when needed. Unlike RSB-Math, these questions require extracting quantities from natural-language context and composing multiple relations before choosing an answer. This subset therefore tests whether a VLA policy can follow multi-sentence compositional problem semantics before acting.

\textbf{(iii) RSB-General} covers non-mathematical semantic understanding, including commonsense QA about everyday tools, locations, household functions, and MMLU-derived multiple-choice questions \cite{MMLU_21}. Together, the three subsets probe semantic grounding in action prediction across controlled calculation, word-problem understanding, and general knowledge.

These sources are not meant to exceed modern language backbones: Qwen3-4B reports over 85\% accuracy on GSM8K and over 70\% on MMLU~\citep{Qwen3TechnicalReport}. Thus, \benchshort{} tests whether semantic competence expected from pretrained backbones is grounded in action prediction after robot fine-tuning.

\subsection{Choice Suites and Visual Grounding}
\label{sec:choice-suites}

Each subset is instantiated in four-choice and ten-choice suites. The four-choice suite uses $\{A,B,C,D\}$; the ten-choice suite uses $\{A,B,C,D,E,F,G,H,I,K\}$, skipping $J$ to avoid visual ambiguity with $I$. The larger suite expands the semantic action space and reduces success by guessing.

The ten-choice suite uses same-color letter blocks with black procedural strokes, making identity visible while reducing color shortcuts. Each episode randomizes block layout and option-to-letter assignment, so the target is determined jointly by the question, options, and mapping rather than a fixed object or position.

\subsection{Instruction Generation and Leakage Control}
\label{sec:instruction-generation}

Instructions are generated from templates that expose the question and option mappings but never the correct answer. Four-choice templates use placeholders such as $\{Q\}$ and $\{\mathrm{MAPA}\},\ldots,\{\mathrm{MAPD}\}$, while ten-choice templates concatenate all mappings into a compact $\{\mathrm{OPTIONS}\}$ field. Seen and unseen template variants test dependence on narrow surface forms.

Thus, the policy must interpret the question, identify the correct option value, and map that option to the visible block. Correct answers are stored only as evaluation metadata, preventing label leakage while enabling post-hoc diagnosis of semantic versus physical failures.

\subsection{Expert Demonstrations and Simulation Evaluation}
\label{sec:eval-metrics}

We instantiate \benchshort{} in a physics-based tabletop simulator with an Aloha-AgileX dual-arm embodiment, multi-view RGB cameras, wrist cameras, robot proprioception. The simulator randomizes candidate-block layout and option-to-block mapping while keeping the answer-selection primitive fixed, so the target is determined by the instruction rather than by position or appearance.

For demonstrations, a scripted expert maps the ground-truth answer to the visible block, chooses an arm by target position, and executes a motion-planned pick-and-place trajectory. Cartesian targets are converted into joint-space trajectories using MPLib with motion planning. Successful seeds are replayed to record trajectories.

At evaluation time, the policy receives only the scene observation and generated instruction, then predicts actions until termination or the step limit. The benchmark logs task success, grasp success, and semantic metadata for diagnosis; the exact metric definitions are given in Section~\ref{sec:exp-metrics}.
This logging is essential because \benchshort{} distinguishes two qualitatively different failures. If a policy never grasps a candidate block, the failure is mainly low-level control. If it grasps a candidate block but not the correct one, the policy has learned the motor primitive but not the semantic target selection. The latter case is the central semantic grounding failure studied in this paper.

\section{Experiments}
\label{sec:experiments}

\begin{table*}[t]
\centering
\small
\resizebox{\textwidth}{!}{%
\rowcolors{4}{TableStripe}{white}
\begin{tabular}{lcccccccccc}
\toprule
Model & Steps & Batch size &
RSB-Math-4 & RSB-Math-10 &
RSB-HardMath-4 & RSB-HardMath-10 &
RSB-General-4 & RSB-General-10 & \textbf{Avg} \\
\midrule
OpenVLA-OFT & 100,000 & 64 & 10.3 & 3.5 & 20.5 & 7.6 & 16.7 & 8.0 & \textbf{11.1} \\
GO1 & 100,000 & 16 & 3.8 & 2.2 & 1.4 & 2.0 & 0.2 & 2.4 & \textbf{2.0} \\
DexVLA & 100,000 & 64 & 13.6 & 2.1 & 5.3 & 1.8 & 12.2 & 3.7 & \textbf{6.5} \\
TinyVLA & 100,000 & 64 & 7.9 & 4.3 & 11.9 & 5.7 & 14.8 & 6.8 & \textbf{8.6} \\
PD-VLA & 100,000 & 64 & 10.9 & 5.3 & 9.8 & 5.1 & 14.2 & 8.8 & \textbf{9.0} \\
$\pi_0$ & 100,000 & 64 & 13 & 5.4 & 25.8 & 6.6 & 18.0 & 7.6 & \textbf{12.7} \\
$\pi_{0.5}$ & 100,000 & 64 & 32.8 & 12.0 & 25.8 & 16.2 & 24.2 & 19.6 & \textbf{21.8} \\
GR00T N1.7 & 100,000 & 64 & 13.8 & 3.4 & 18.4 & 9.2 & 23.4 & 7.4 & \textbf{12.6} \\
QwenGR00T & 100,000 & 64 & 18.4 & 3.4 & 24.4 & 1.8 & 15.8 & 0.6 & \textbf{10.7} \\
\bottomrule
\end{tabular}
\rowcolors{1}{}{}
}
\caption{Main evaluation results in \textbf{Task Success Rate (TSR, \%)} after fine-tuning on expert demonstrations. Avg is computed over the six evaluation suites when all six are available. Full GSR/TSR decomposition is provided in Table~\ref{tab:full-main-results} in the appendix.}
\label{tab:main-results}
\end{table*}

\begin{table*}[t]
\centering
\small
\resizebox{\textwidth}{!}{%
\rowcolors{4}{TableStripe}{white}
\begin{tabular}{lccccccc}
\toprule
Model &
RSB-Math-4 & RSB-Math-10 &
RSB-HardMath-4 & RSB-HardMath-10 &
RSB-General-4 & RSB-General-10 & \textbf{Avg} \\
\midrule
OpenVLA-OFT & -19.2 & -6.9 & -6.0 & -0.9 & -7.9 & -2.2 & \textbf{-7.2} \\
GO1 & -27.8 & -8.6 & -30.1 & -8.8 & -33.0 & -8.4 & \textbf{-19.4} \\
DexVLA & -4.8 & 2.2 & -5.1 & -2.3 & -14.6 & 4.3 & \textbf{-3.4} \\
TinyVLA & -4.4 & 0.1 & 0.6 & -0.7 & 1.7 & 4.1 & \textbf{0.2} \\
PD-VLA & 1.2 & 2.2 & -2.1 & 4.3 & 14.0 & -0.5 & \textbf{3.2} \\
$\pi_0$ & -14.3 & -5.1 & 1.1 & -3.8 & -9.2 & -2.7 & \textbf{-5.7} \\
$\pi_{0.5}$ & 11.3 & 2.2 & 1.1 & 6.9 & -1.1 & 10.7 & \textbf{5.2} \\
GR00T N1.7 & -14.3 & -7.2 & -8.6 & -0.7 & -1.6 & -2.8 & \textbf{-5.9} \\
QwenGR00T & -8.0 & -7.2 & -0.7 & -9.1 & -7.4 & -10.4 & \textbf{-7.1} \\
\bottomrule
\end{tabular}
\rowcolors{1}{}{}
}
\caption{\textbf{Normalized Semantic Grounding score (nSG, \%)} for evaluated VLA models. nSG factors out grasp success by measuring semantic target selection conditioned on grasping a candidate block; 0 corresponds to random target selection and negative values indicate worse-than-random selection. Avg is computed over the six evaluation suites when all six are available.}
\label{tab:nsg-results}
\end{table*}

\subsection{Experimental Protocol}
\label{sec:exp-protocol}

All models follow the same train--test protocol. For each \benchshort{} suite in Table~\ref{tab:dataset-stats}, we collect expert demonstrations from the training split, fine-tune the policy, and evaluate it on held-out semantic questions. Training and evaluation questions are disjoint to prevent memorizing question-answer pairs.

The number of training questions is determined by the source and cost of each semantic subset. RSB-Math uses 500 procedurally generated arithmetic questions, which already cover the controlled operator and distractor patterns while keeping expert trajectory generation lightweight. RSB-HardMath uses the full 7,473-question GSM8K training split, and RSB-General uses 10,000 sampled MMLU-style questions to provide broad commonsense and factual coverage. For each subset, the same question pool is used for both the 4-choice and 10-choice suites so that choice-set size changes while the underlying semantic distribution remains comparable.

To make comparison fair, we use comparable robot fine-tuning budgets, measured by training steps times batch size, whenever supported. Each model is reproduced and fine-tuned with its official codebase, using repository defaults unless otherwise stated.

At evaluation time, the policy receives only the observation and generated instruction; answer labels are never provided. For each model and suite, we run 500 simulation episodes and report average success rates. We release the full training data and simulation evaluation code for every \benchshort{} suite.

\begin{table}[t]
\centering
\small
\begin{tabular}{llcc}
\toprule
Subset & Source & Choices & Train Questions \\
\midrule
RSB-Math-4 & easy arithmetic & 4 & 500 \\
RSB-Math-10 & easy arithmetic & 10 & 500 \\
RSB-HardMath-4 & GSM8K & 4 & 7,473 \\
RSB-HardMath-10 & GSM8K & 10 & 7,473 \\
RSB-General-4 & MMLU-style QA & 4 & 10,000 \\
RSB-General-10 & MMLU-style QA & 10 & 10,000 \\
\bottomrule
\end{tabular}
\caption{Training-set statistics for each evaluation suite. Training and evaluation questions are disjoint within each subset.}
\label{tab:dataset-stats}
\end{table}

\subsection{Evaluated Models}
\label{sec:exp-models}

We select evaluated models based on representativeness and reproducibility. Specifically, we evaluate recent VLA models including GO1~\citep{GO1_25_blog}, OpenVLA~\citep{OpenVLA_24}, DexVLA~\citep{DexVLA_25_arxiv}, TinyVLA~\citep{TinyVLA_25_arxiv}, PD-VLA~\citep{LLaVA-VLA_25_arxiv}, $\pi_0$~\citep{PI0}, $\pi_{0.5}$~\citep{PI05_25}, GR00T N1.7~\citep{GR00T_25}, and QwenGR00T~\citep{starVLA_26_arxiv}. Table~\ref{tab:main-results} reports fine-tuning configurations and results; Appendix~\ref{sec:appendix-training-details} gives model-specific details.

\subsection{Metrics}
\label{sec:exp-metrics}

We report three metrics for each evaluation suite. \textbf{Task Success Rate (TSR)} measures the fraction of episodes in which the policy grasps the correct answer block specified by the semantic question and option mapping. \textbf{Grasp Success Rate (GSR)} measures the fraction of episodes in which the policy grasps any candidate answer block, regardless of correctness. Appendix~\ref{sec:appendix-grasp-success-criteria} provides the detailed criteria for grasp success. To factor out differences in low-level grasping ability, we further define a \textbf{normalized Semantic Grounding score (nSG)}:
\begin{equation}
    \mathrm{nSG}
    =
    \frac{\mathrm{TSR}/\mathrm{GSR} - 1/N}{1 - 1/N},
\end{equation}
where $N$ is the number of candidate choices. This score measures whether a model selects the semantically correct target conditioned on successfully grasping a candidate block: $\mathrm{nSG}=0$ corresponds to random target selection, while $\mathrm{nSG}=1$ corresponds to perfect semantic target selection among successful grasps.

\subsection{Main Results}
\label{sec:main-results}

Table~\ref{tab:main-results} reports TSR across all suites, and Table~\ref{tab:nsg-results} reports nSG after normalizing for grasp success. If a policy learns the manipulation primitive without reliable semantic understanding, TSR should remain low and nSG should stay near or below zero, especially in harder semantic domains and larger choice sets.

Appendix Table~\ref{tab:full-main-results} provides the full GSR--TSR decomposition. High GSR with low TSR means the policy can grasp answer blocks but fails to select the semantically correct one; nSG asks whether this target selection is better than random among successful grasps.

Overall, most VLA models behave close to random target selection once low-level grasping is factored out: 25\% for four-choice suites and 10\% for ten-choice suites. Most average nSG scores are near or below zero, showing that successful grasps are not consistently guided by the semantic answer. This is not simply a failure to move the robot: several models achieve high GSR in the full decomposition, but their TSR remains low because they often grasp the wrong candidate. The gap is especially visible in the 10-choice suites, where the action space contains more plausible targets and shortcut-based selection becomes less reliable.

The main exception is $\pi_{0.5}$, which achieves the highest average TSR and the only clearly positive average nSG. A plausible explanation is that $\pi_{0.5}$ uses subtask annotations during robot-data pretraining, which may provide weak supervision for decomposing high-level instructions into intermediate semantic decisions and then following those decisions during action generation. Even so, its nSG remains modest, indicating that current VLA training is far from robust semantic grounding in action prediction.

\subsection{Beyond Blocks: Everyday Object Targets}
\label{sec:beyond-blocks}

\begin{figure}[h]
  \centering
  \includegraphics[width=\columnwidth]{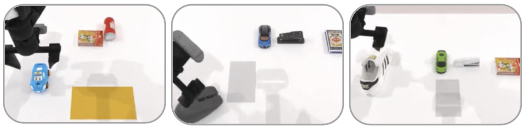}
  \caption{Examples of the \emph{Beyond Blocks} setting, where candidate answer targets are replaced with everyday objects instead of uniform lettered blocks.}
  \label{fig:beyond_blocks}
\end{figure}

\begin{figure*}[ht]
  \centering
  \includegraphics[width=\textwidth]{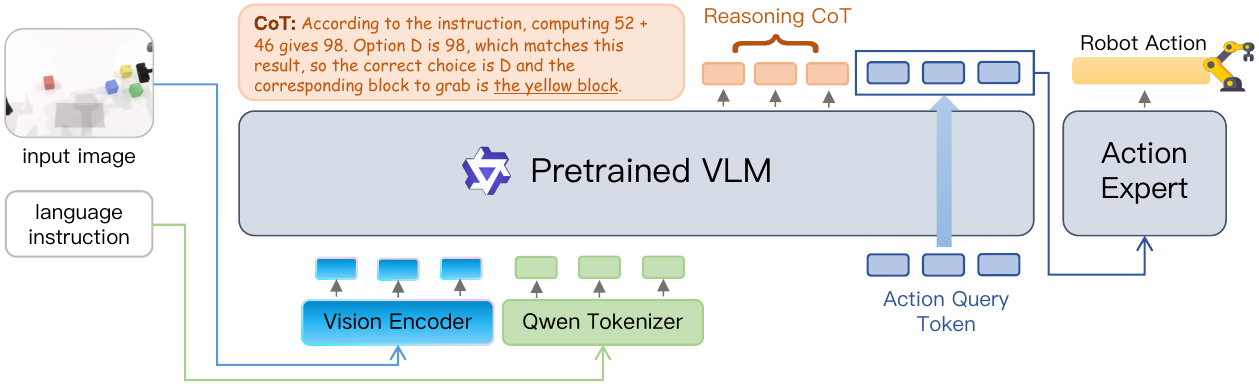}
  \caption{Overview of \textbf{ReasoningVLA}. The VLM generates a textual CoT, then uses Action Query Tokens to pass semantic context to the Action Expert for action-chunk generation.}
  \label{fig:reasoning_vla}
\end{figure*}

The main suites use lettered blocks to isolate semantic target selection from object-specific grasping difficulty. To test whether this proxy causes the observed failures, we replace candidate blocks with everyday objects such as toy cars, playing cards, and shoes, and build three scenes matching the three \benchshort{} semantic levels. Using the same GSR/TSR protocol, Appendix~\ref{sec:appendix-beyond-blocks} shows that grasp success remains high while TSR stays low. Thus, the benchmark difficulty mainly comes from semantic grounding in action prediction rather than from the block interface, making blocks a controlled and sufficient proxy.

\section{Failed Exploration}
\label{sec:failed-exploration}

Beyond benchmarking existing VLA models, we explored two natural interventions for reducing the semantic grounding gap. One makes the VLA verbalize an intermediate semantic solution before predicting actions; the other adds language-centric supervision during robot fine-tuning. Both are negative results: neither reliably makes pretrained semantic understanding participate in action prediction.

\newcommand{\gain}{\textcolor{green!50!black}{$\uparrow$}}
\newcommand{\drop}{\textcolor{red!75!black}{$\downarrow$}}

\begin{table*}[t]
\centering
\small
\resizebox{\textwidth}{!}{%
\begin{tabular}{lccccccc}
\toprule
Method &
RSB-Math-4 & RSB-Math-10 &
RSB-HardMath-4 & RSB-HardMath-10 &
RSB-General-4 & RSB-General-10 &
TSR Avg \\
\midrule
QwenGR00T (baseline) & 18.4 & 3.4 & 24.4 & 1.8 & 15.8 & 0.6 & 10.7 \\
ReasoningVLA & 27.0 \gain & 7.3 \gain & 28.6 \gain & 5.8 \gain & 20.6 \gain & 6.8 \gain & 16.0 \gain \\
QwenGR00T (Cotrain) & 13.2 \drop & 2.4 \drop & 20.2 \drop & 0.8 \drop & 12.2 \drop & 0.2 \drop & 8.2 \drop \\
\bottomrule
\end{tabular}
}
\caption{TSR results for failed exploration attempts compared with the QwenGR00T baseline. ReasoningVLA improves TSR but remains far from reliable semantic grounding in action prediction, while VLA cotraining consistently reduces TSR. TSR Avg is the mean over the six evaluation suites.}
\label{tab:reasoning-vla-results}
\label{tab:cotrain-results}
\end{table*}

\subsection{Exploration 1: ReasoningVLA}

ReasoningVLA makes the semantic decision explicit before action generation. It trains the model to produce a textual CoT identifying the target option block, then conditions action prediction on this intermediate solution. This tests a natural hypothesis: if the VLM backbone can first solve the semantic question in language space, the downstream action module may have an easier target-selection problem. The risk is that the generated answer must still be bound to the current scene and preserved through continuous action generation.

\textbf{Architecture.}
ReasoningVLA uses a Qwen3-VL-4B backbone and a DiT-based Action Expert, as shown in Figure~\ref{fig:reasoning_vla}. The VLM autoregressively generates a CoT between \texttt{<think>} and \texttt{</think>}, then appends eight \texttt{<|action|>} Action Query Tokens. Their last-layer hidden states are passed to the Action Expert through cross-attention, and the Action Expert generates an action chunk with flow matching. The Action Query Tokens therefore serve as the interface between the textual semantic solution and the continuous action generator.

CoT supervision is distilled from Gemini 3 Flash~\citep{gemini3_blog} and attached to each robot demonstration; Appendix~\ref{sec:appendix-reasoningvla-data} provides the prompt format and examples.

\textbf{Training Objective.}
ReasoningVLA optimizes a weighted action-generation and CoT-generation objective:
\begin{equation}
    \mathcal{L} = 0.9\,\mathcal{L}_{\mathrm{FM}} + 0.1\,\mathcal{L}_{\mathrm{CoT}}.
\end{equation}
Here $\mathcal{L}_{\mathrm{FM}}$ supervises the action chunk and $\mathcal{L}_{\mathrm{CoT}}$ supervises next-token prediction for the distilled CoT.

Table~\ref{tab:reasoning-vla-results} shows that ReasoningVLA improves average TSR over QwenGR00T, but its GSR decreases and the absolute TSR remains low, especially in 10-choice suites. This suggests that CoT supervision can help recover part of the semantic target selection ability, but it does not reliably ground the selected answer into robot actions. Explicit reasoning traces alone are therefore insufficient for semantic grounding in action prediction.

\subsection{Exploration 2: VLA Cotrain}
The second attempt is to preserve the VLM backbone's semantic competence by adding VQA supervision during robot fine-tuning. As shown in Table~\ref{tab:cotrain-results}, this cotraining strategy does not improve semantic grounding in action prediction: TSR drops on all six \benchshort{} suites, and the average TSR decreases from 10.7\% to 8.2\%. This suggests that language-centric auxiliary supervision may conflict with the action-learning objective and perturb representations needed for VLA adaptation, consistent with the finding of VLM4VLA~\citep{VLM4VLA_25}. Appendix~\ref{sec:appendix-vla-cotrain-details} provides the training details.

\section{Error Analysis}
\label{sec:error-analysis}

We analyze RSB-Math episodes where the policy grasps a candidate block but still fails the task. Since the robot has interacted with an answer block, these errors directly reveal whether the selected target follows the instruction semantics rather than whether the robot can execute a grasp.

\begin{table}[h]
\centering
\small
\resizebox{\columnwidth}{!}{%
\begin{tabular}{lcc}
\toprule
Error type & ReasoningVLA & QwenGR00T \\
\midrule
Grasped but not placed correctly & 4.36 & 4.08 \\
Incorrect CoT, wrong target & 6.70 & -- \\
Correct CoT, wrong target & 89.93 & -- \\
Wrong target after grasp & -- & 95.92 \\
\bottomrule
\end{tabular}
}
\caption{Error analysis on RSB-Math for episodes with grasp success but task failure. Values are percentages within this failure subset.}
\label{tab:error-analysis}
\end{table}

Table~\ref{tab:error-analysis} shows that most failures are target-selection errors rather than placement failures. QwenGR00T chooses the wrong target in 95.92\% of grasp-success/task-failure cases, and ReasoningVLA shows the same pattern. This supports the central diagnosis of \benchshort{}: current VLA models can often learn the answer-block manipulation primitive, but they still do not know which block should be selected from the instruction.

ReasoningVLA further exposes a \emph{reasoning-following} failure. Only 6.70\% of its grasp-success/task-failure episodes are associated with an incorrect CoT; most occur when the CoT identifies the correct answer but the action still grasps the wrong block. Thus, improving language-space semantic decisions is not sufficient unless the action pathway reliably follows those decisions.

\section{Discussion}
\label{sec:discussion}

\textbf{What does RSB diagnose?}
\benchshort{} evaluates semantic grounding in action prediction rather than standalone question answering. It asks whether a VLA model can use instruction semantics to choose the correct physical target during action prediction, after accounting for its ability to grasp candidate objects. This distinction is important: a VLA may contain a capable pretrained backbone, yet still act as if it does not understand the instruction if the action pathway relies on imitation shortcuts, visual priors, or poorly routed semantic features. The GSR--TSR decomposition and nSG score make this gap visible by separating grasping ability from semantic target selection.

\textbf{Implications for VLA training.}
The results suggest that simply attaching a strong VLM to an action expert is not enough to obtain semantically grounded action prediction. A more promising direction may require training objectives and interfaces that explicitly preserve the selected semantic target and expose it to the action module in a stable, scene-grounded form. In this sense, \benchshort{} provides not only an evaluation suite, but also a diagnostic target for future VLA architectures: successful models should maintain high grasp success while raising TSR and nSG far above random selection.

\section{Conclusion}
\label{sec:conclusion}

We presented \methodname{} (\benchshort{}), a benchmark for diagnosing semantic grounding in VLA action prediction. By converting math, hard-math, and general-knowledge questions into embodied answer-selection tasks, \benchshort{} makes the correct action depend on instruction understanding rather than visual or action-distribution shortcuts. Across representative VLA models, many policies learn the grasping primitive, but their target choices remain near or below random once grasp success is controlled for. This reveals that current VLA training often fails to route semantic decisions from the pretrained backbone into the action pathway, and motivates future VLA systems whose actions are genuinely grounded in instruction semantics.

\bibliography{custom}

\newpage

\appendix

\section{Full Main Evaluation Results}
\label{sec:appendix-full-main-results}

Table~\ref{tab:full-main-results} reports the full GSR/TSR decomposition for all evaluated models. The Avg columns summarize the mean GSR and TSR over the six \benchshort{} evaluation suites when all six results are available.

\begin{table*}[t]
\centering
\small
\resizebox{\textwidth}{!}{%
\rowcolors{4}{TableStripe}{white}
\begin{tabular}{lcccccccccccccccc}
\toprule
\multirow{2}{*}{Model} & \multirow{2}{*}{Steps} & \multirow{2}{*}{Batch size} &
\multicolumn{2}{c}{RSB-Math-4} &
\multicolumn{2}{c}{RSB-Math-10} &
\multicolumn{2}{c}{RSB-HardMath-4} &
\multicolumn{2}{c}{RSB-HardMath-10} &
\multicolumn{2}{c}{RSB-General-4} &
\multicolumn{2}{c}{RSB-General-10} &
\multicolumn{2}{c}{\textbf{Avg}} \\
\cmidrule(lr){4-5}
\cmidrule(lr){6-7}
\cmidrule(lr){8-9}
\cmidrule(lr){10-11}
\cmidrule(lr){12-13}
\cmidrule(lr){14-15}
\cmidrule(lr){16-17}
& & & GSR & TSR & GSR & TSR & GSR & TSR & GSR & TSR & GSR & TSR & GSR & TSR & GSR & TSR \\
\midrule
OpenVLA-OFT & 100,000 & 64 & 97 & 10.3 & 92.3 & 3.5 & 100.0 & 20.5 & 83.1 & 7.6 & 87.5 & 16.7 & 100.0 & 8.0 & \textbf{93.3} & \textbf{11.1} \\
GO1 & 100,000 & 16 & 91.2 & 3.8 & 96.0 & 2.2 & 58.4 & 1.4 & 96.8 & 2.0  & 70.8 & 0.2 & 100.0 & 2.4 & \textbf{85.5} &  \textbf{2.0} \\
DexVLA & 100,000 & 64 & 63.6 & 13.6 & 17.5 & 2.1 & 25.0 & 5.3 & 22.7 & 1.8 & 87.0 & 12.2 & 26.7 & 3.7 & \textbf{40.4} & \textbf{6.5} \\
TinyVLA & 100,000 & 64 & 36.4 & 7.9 & 42.8 & 4.3 & 46.8 & 11.9 & 61.0 & 5.7 & 56.3 & 14.8 & 49.6 & 6.8 & \textbf{48.8} & \textbf{8.6} \\
PD-VLA & 100,000 & 64 & 42.1 & 10.9 & 44.2 & 5.3 & 41.8 & 9.8 & 36.8 & 5.1 & 40.0 & 14.2 & 92.3 & 8.8 & \textbf{49.5} & \textbf{9.0} \\
$\pi_0$ & 100,000 & 64 & 91.0 & 13.0 & 100.0 & 5.4 & 100.0 & 25.8 & 100.0 & 6.6 & 99.4 & 18.0 & 100.0 & 7.6 & \textbf{98.4} & \textbf{12.7} \\
$\pi_{0.5}$ & 100,000 & 64 & 98.0 & 32.8 & 100.0 & 12.0 & 100.0 & 25.8 & 100 & 16.2 & 100.0 & 24.2 & 100.0 & 19.6 & \textbf{99.7} & \textbf{21.8} \\
GR00T N1.7 & 100,000 & 64 & 96.6 & 13.8 & 97.6 & 3.4 & 99.2 & 18.4 & 98.0 & 9.2 & 98.2 & 23.4 & 99.2 & 7.4 & \textbf{98.1} & \textbf{12.6} \\
QwenGR00T & 100,000 & 64 & 96.8 & 18.4  & 97.6 & 3.4 & 99.6 & 24.4 & 98.8 & 1.8 & 81.2 & 15.8 & 96.8 & 0.6 & \textbf{95.1} & \textbf{10.7} \\
\bottomrule
\end{tabular}
\rowcolors{1}{}{}
}
\caption{Full GSR/TSR decomposition for the main evaluation. GSR measures grasping any candidate block, while TSR requires grasping the correct answer block. Avg reports the mean GSR and TSR over all six evaluation suites when all six are available. All success rates are percentages and higher is better.}
\label{tab:full-main-results}
\end{table*}

\section{Beyond Blocks Results}
\label{sec:appendix-beyond-blocks}

Tables~\ref{tab:beyond-blocks-pi05-results} and~\ref{tab:beyond-blocks-groot-results} report the Beyond Blocks results for $\pi_{0.5}$ and GR00T N1.7. After replacing lettered blocks with everyday objects, both models show evaluation trends consistent with the block-based setting: GSR remains high, while TSR remains much lower than grasp success. This indicates that the main bottleneck is still semantic target selection rather than object-specific grasping difficulty. Therefore, the block-based \benchshort{} task is sufficient as a controlled proxy for evaluating semantic grounding in current VLA action prediction.

\begin{table}[h]
\centering
\small
\begin{tabular}{lcc}
\toprule
Scene & GSR & TSR \\
\midrule
RSB-Math-4  & 96.2 & 32.4 \\
RSB-Math-10 & 96.8 & 11.8 \\
RSB-HardMath-4 & 97.4 & 24.6 \\
RSB-HardMath-10  & 96.0 & 15.8 \\
RSB-General-4 & 97.4 & 23.6 \\
RSB-General-10 & 97.2 & 19.2 \\
\midrule
\textbf{Avg} & \textbf{96.8} & \textbf{21.2} \\
\bottomrule
\end{tabular}
\caption{Beyond Blocks evaluation results for $\pi_{0.5}$. GSR measures grasping any candidate object, while TSR requires grasping the object associated with the correct semantic answer.}
\label{tab:beyond-blocks-pi05-results}
\end{table}

\begin{table}[h]
\centering
\small
\begin{tabular}{lcc}
\toprule
Scene & GSR & TSR \\
\midrule
RSB-Math-4 & 95.2 & 13.4 \\
RSB-Math-10 & 96.4 & 3.2 \\
RSB-HardMath-4 & 98.4 & 18.8 \\
RSB-HardMath-10 & 97.8 & 9.4 \\
RSB-General-4 & 96.8 & 22.6 \\
RSB-General-10 & 97.2 & 7.0 \\
\midrule
\textbf{Avg} & \textbf{97.0} & \textbf{12.4} \\
\bottomrule
\end{tabular}
\caption{Beyond Blocks evaluation results for GR00T N1.7. GSR measures grasping any candidate object, while TSR requires grasping the object associated with the correct semantic answer.}
\label{tab:beyond-blocks-groot-results}
\end{table}

\section{Grasp Success Criteria}
\label{sec:appendix-grasp-success-criteria}

\paragraph{Grasp success detection.}
GSR measures the percentage of episodes in which the policy successfully grasps any candidate answer block, regardless of whether the grasped block corresponds to the correct answer. A grasp is counted only when all of the following conditions are satisfied: (1) one gripper is in contact with a candidate block; (2) at least one of the left or right grippers is closed; and (3) the contacted block is either lifted above the tabletop or remains in stable gripper contact for at least eight consecutive simulation steps. This definition avoids counting accidental gripper closure, empty grasps, or contacts with non-candidate objects as grasp success. A policy that grasps the wrong candidate block is therefore counted as successful under GSR but unsuccessful under TSR, which is what allows the GSR--TSR gap to diagnose semantic target-selection failures.

\paragraph{Normalized semantic grounding score.}
The nSG score is computed only when GSR is non-zero and should be interpreted together with GSR. It measures semantic target selection conditioned on the policy having successfully grasped a candidate block. Because the score subtracts the random-selection baseline $1/N$, values near zero indicate random semantic selection among grasped candidates, positive values indicate better-than-random semantic grounding, and negative values indicate worse-than-random target selection.

\section{ReasoningVLA Data Construction}
\label{sec:appendix-reasoningvla-data}

\paragraph{CoT annotation source.}
Default VLA demonstrations contain observations, language instructions, and expert action chunks, but they do not include textual explanations for the semantic decision behind each action. We therefore augment the \benchshort{} training demonstrations with CoT annotations distilled from Gemini 3 Flash~\citep{gemini3_blog}. For each training instruction, the teacher model receives the multiple-choice question and options, solves the semantic problem, and outputs a short rationale that includes the expression, computed answer, matched option, and corresponding color block.

\paragraph{Prompt format.}
We use a fixed system prompt to constrain the teacher to produce concise annotations and to follow a deterministic option-to-color mapping:
\begin{promptbox}[title=System Prompt]
\small
\ttfamily
You are a data annotation assistant for robot block-picking tasks. Given an instruction, generate a concise solution rationale. Only output one paragraph. Do not output JSON or extra explanations. The reasoning must include the math expression, computed result, matched option, and corresponding color block. The fixed color mapping is: A=red block, B=green block, C=blue block, D=yellow block.
\end{promptbox}
The resulting rationale is then wrapped with \texttt{<think>} and \texttt{</think>} before being used as CoT supervision. We discard teacher responses that do not contain a unique final option or that are inconsistent with the ground-truth answer metadata.

\paragraph{Example annotation.}
For an instruction such as ``what is 27 minus 17? options: (A) 4 (B) 10 (C) 11 (D) 14. place the correct option block in the gray answer zone,'' the distilled annotation is:
\begin{promptbox}[title=Example CoT Annotation]
\small
\texttt{<think>} 27 minus 17 equals 10. Among the options, option B is 10. According to the fixed color mapping, option B corresponds to the green block, so the robot should grasp the green block. \texttt{</think>}
\end{promptbox}
This CoT is concatenated with the original observation--instruction demonstration. The action target remains the same expert action chunk, so the augmented sample supervises both the language-space semantic solution and the continuous action-generation pathway.

\section{VLA Cotraining Details}
\label{sec:appendix-vla-cotrain-details}

\paragraph{Motivation.}
The cotraining experiment tests whether language-centric supervision can preserve the VLM backbone's semantic competence during robot fine-tuning. The baseline QwenGR00T is trained only on \benchshort{} robot demonstrations, where each sample contains observations, a semantic instruction, and an expert action chunk. The cotraining variant keeps the same robot demonstration data and evaluation protocol, but additionally injects visual question answering (VQA) samples from RoboVQA~\citep{RoboVQA_23_arxiv} during fine-tuning so that the shared VLM backbone is optimized for both semantic question answering and action prediction.

\paragraph{Training mixture.}
We implement cotraining based on the starVLA framework. Mixed fine-tuning batches are constructed from two sources: robot-demonstration samples use the standard VLA input format and supervise the action-generation pathway with expert action chunks, while VQA samples use image--question--answer pairs and supervise the language modeling pathway with next-token prediction over the answer text. The two data streams share the same VLM backbone, while the action expert is updated only by robot-demonstration samples. This design is intended to regularize the Semantic Expert without changing the robot task definition or giving the policy access to evaluation answers.

\paragraph{Training setup.}
We train the cotraining model for 100{,}000 steps on 8 NVIDIA H100 GPUs using DeepSpeed ZeRO-2 optimization. The global batch size is 64 for the VLA robot-demonstration stream and 32 for the VQA stream.

\paragraph{Objective.}
The cotraining objective combines the original VLA action loss with a language modeling loss on VQA answers:
\begin{equation}
    \mathcal{L}_{\mathrm{cotrain}}
    =
    \mathcal{L}_{\mathrm{action}}
    +
    0.1\,\mathcal{L}_{\mathrm{VQA}},
\end{equation}
where $\mathcal{L}_{\mathrm{action}}$ is the action-generation loss used by the QwenGR00T baseline and $\mathcal{L}_{\mathrm{VQA}}$ is a next-token prediction loss for RoboVQA responses. The VQA branch therefore encourages the backbone to retain language-space semantic answering ability, while the action branch continues to optimize imitation learning on expert robot trajectories.

\paragraph{Evaluation.}
After cotraining, the model is evaluated exactly like the baseline: it receives only the simulator observation and the generated \benchshort{} instruction, and must output robot actions without access to the correct answer label or VQA supervision. This makes the comparison in Table~\ref{tab:cotrain-results} a direct test of whether preserving language-oriented supervision during fine-tuning improves semantic grounding in action prediction.

\section{Training Details for Evaluated Models}
\label{sec:appendix-training-details}

All evaluated models are trained on 8 NVIDIA H100 GPUs. To make the training budget comparable across models, we keep the total number of optimization samples fixed by matching \emph{training steps} $\times$ \emph{global batch size}. Other model-specific training details, including optimizer settings, learning-rate schedules, precision settings, and architecture-specific data formatting, follow the default configurations of the corresponding official codebases whenever possible.

\paragraph{GO1.}
GO1 is an open generalist VLA robotic foundation model released by OpenDriveLab, designed as a scalable policy for language-conditioned robot control~\citep{GO1_25_blog}. We reproduce GO1 with its official codebase and fine-tune it on \benchshort{} demonstrations using the observation, instruction, and action interface expected by the released implementation.

\paragraph{OpenVLA.}
OpenVLA is an open-source VLA model that adapts a pretrained vision-language backbone into an autoregressive robot policy, representing robot actions as tokens predicted from visual observations and language instructions~\citep{OpenVLA_24}. We fine-tune OpenVLA from its pretrained checkpoint on the \benchshort{} training split and evaluate the resulting policy with the same simulator protocol as the other models.

\paragraph{DexVLA.}
DexVLA augments a VLM backbone with a plug-in diffusion expert, using the VLM for semantic perception and instruction processing while delegating continuous robot control to a diffusion-based action module~\citep{DexVLA_25_arxiv}. We fine-tune DexVLA on the same semantic answer-selection demonstrations used for the other evaluated models.

\paragraph{TinyVLA.}
TinyVLA is a compact and data-efficient VLA architecture designed to reduce inference cost while maintaining robot manipulation performance~\citep{TinyVLA_25_arxiv}. We train TinyVLA on the \benchshort{} expert demonstrations as a small VLA baseline under the same train--test split.

\paragraph{PD-VLA.}
PD-VLA targets VLA models with action chunking and accelerates autoregressive action decoding through parallel fixed-point decoding, preserving the underlying action-chunking policy interface while improving inference efficiency~\citep{LLaVA-VLA_25_arxiv}. We fine-tune and evaluate the released PD-VLA-style implementation on the \benchshort{} training and evaluation suites.

\paragraph{$\pi_0$ and $\pi_{0.5}$.}
$\pi_0$ is a generalist VLA flow model built around a pretrained PaliGemma-style VLM backbone and an action expert, using a mixture-of-transformers style interface to connect semantic processing with continuous action generation via flow matching~\citep{PI0}. $\pi_{0.5}$ extends this family toward open-world generalization and uses additional robot-data pretraining signals, including subtask-style supervision, that can help decompose instructions before action prediction~\citep{PI05_25}. We fine-tune both models on the \benchshort{} expert demonstrations with their official training pipelines.

\paragraph{GR00T N1.7.}
GR00T N1 is an open foundation model for generalist humanoid robots, combining multimodal instruction understanding with robot action generation for whole-body or manipulation-oriented control~\citep{GR00T_25}. We fine-tune GR00T N1.7 on the corresponding \benchshort{} training split and evaluate it using the same 500-episode simulator protocol.

\paragraph{QwenGR00T.}
QwenGR00T is a Qwen-backed VLA implementation built within the StarVLA codebase, which provides a modular framework for constructing VLA policies from interchangeable VLM backbones and action experts~\citep{starVLA_26_arxiv}. We fine-tune QwenGR00T on the \benchshort{} demonstrations using the same train--test split as the other VLA models.

\end{document}